\documentclass[journal]{IEEEtran}
\usepackage{booktabs}
\usepackage{multirow}
\usepackage{graphicx}
\usepackage[table,xcdraw]{xcolor}
\usepackage[normalem]{ulem}
\useunder{\uline}{\ul}{}
\usepackage{amsmath}
\usepackage{amsfonts}
\DeclareMathOperator*{\argmax}{arg\,max}

\usepackage{amsthm}
\theoremstyle{definition}
\newtheorem{definition}{Definition}[section]
\usepackage{caption}
\usepackage{subcaption}

\ifCLASSINFOpdf
\else
\fi
\begin{document}
\title{Learning Graph Structures with Transformer for Multivariate Time Series Anomaly Detection in IoT}

\author{Zekai~Chen,~\IEEEmembership{Student Member,~IEEE,}
        Dingshuo~Chen,
        Xiao~Zhang,~\IEEEmembership{Member,~IEEE},
        Zixuan Yuan,
        and~Xiuzhen~Cheng,~\IEEEmembership{Fellow,~IEEE}% <-this % stops a space
\thanks{Z. Chen is with the Department
of Computer Science, George Washington University, Washington,
DC, 20052 USA (email: zech\_chan@gwu.edu)}% <-this % stops a space
\thanks{Z. Yuan is with the School of Business, Rutgers University, New Jersey, 08901 USA (email: zy101@rutgers.edu)}
\thanks{X. Zhang (corresponding author), X. Cheng and D. Chen are with School of Computer Science and Technology, Shandong University, China (emails: xiaozhang@sdu.edu.cn, xzcheng@sdu.edu.cn)}% <-this % stops a space
% \thanks{Manuscript received April 19, 2005; revised August 26, 2015.}
}

\maketitle

\begin{abstract}
Many real-world IoT systems, which include a variety of internet-connected sensory devices, produce substantial amounts of multivariate time series data. Meanwhile, vital IoT infrastructures like smart power grids and water distribution networks are frequently targeted by cyber-attacks, making anomaly detection an important study topic. Modeling such relatedness is, nevertheless, unavoidable for any efficient and effective anomaly detection system, given the intricate topological and nonlinear connections that are originally unknown among sensors. Furthermore, detecting anomalies in multivariate time series is difficult due to their temporal dependency and stochasticity. This paper presented GTA, a new framework for multivariate time series anomaly detection that involves automatically learning a graph structure, graph convolution, and modeling temporal dependency using a Transformer-based architecture. The connection learning policy, which is based on the Gumbel-softmax sampling approach to learn bi-directed links among sensors directly, is at the heart of learning graph structure. To describe the anomaly information flow between network nodes, we introduced a new graph convolution called Influence Propagation convolution. In addition, to tackle the quadratic complexity barrier, we suggested a multi-branch attention mechanism to replace the original multi-head self-attention method. Extensive experiments on four publicly available anomaly detection benchmarks further demonstrate the superiority of our approach over alternative state-of-the-arts. Codes are available at https://github.com/ZEKAICHEN/GTA.
\end{abstract}


% \fi
\IEEEpeerreviewmaketitle

\section{Introduction}
\label{intro}
Due to the fast rising number of Internet-connected sensory devices, the Internet of Things (IoT) infrastructure has created vast sensory data. IoT data is often characterized by its speed in terms of geographical and temporal dependency \cite{Mahdavinejad2018,Cai2019}, and it is frequently subjected to correspondingly rising abnormalities and cyberattacks \cite{Mohammadi2018,Deng2021}.
% With the rapidly growing volume of Internet-connected sensory devices, the Internet of Things (IoT) infrastructure generates massive data characterized by its velocity in terms of spatial and temporal dependency , accompanied by commensurately increasing anomalies and attacks . 
Many critical infrastructures constructed on top of Cyber-Physical Systems (CPS) \cite{Cai2020}, such as smart power grids, water treatment and distribution networks, transportation, and autonomous cars, are especially in need of security monitoring \cite{Li2019,Zheng2020,Deng2021}. As a result, an efficient and accurate anomaly detection system has great research value because it can help with continuous monitoring of fundamental controls or indicators and promptly provide notifications for any probable anomalous occurrence.

In this work, we focus on anomaly detection for multivariate time series \cite{Hundman2018} as a copious amount of IoT sensors in many real-life scenarios consecutively generate substantial volumes of time series data. For instance, in a Secure Water Distribution (WADI) system \cite{Ahmed2017}, multiple sensing measurements such as flowing meter, transmitting level, valve status, water pressure level, etc., are recorded simultaneously at each timestamp to form a multivariate time series. In this case, the central water treatment testbed is also known as an \emph{entity}. It is commonly accepted to detect anomalies from the entity-level instead of the sensor-level since the overall status detection is generally worth more concern and less expensive. Predominantly, data from these sensors are highly correlated in a complex topological and nonlinear fashion: for example, opening a valve would result in pressure and flow rate changes, leading to further chain reactions of other sensors within the same entity following an internal mechanism. Nevertheless, \emph{the dependencies among sensors are initially hidden and somehow costly to access in most real-life scenarios, leading to an intuitive question of how to model such complicated relationships between sensors without knowing prior information?}

Recently, deep learning-based techniques have demonstrated some promising improvements in anomaly detection due to the superiority in sequence modeling over high-dimensional datasets. Generally, the existing approaches can roughly fall into two lines: reconstruction-based models (R-model) \cite{Aggarwal2013,Park2017,Zong2018,Li2019,Su2019} and forecasting-based models (F-model) \cite{Angiulli2002,Lazarevic2005,Hundman2018,Liang2018,Zhao2020,Deng2021}. For example, Auto-Encoders (AE) \cite{Aggarwal2013} is a popular approach for anomaly detection, which uses reconstruction error as an outlier score. More recently, Generative Adversarial Networks (GANs) \cite{Goodfellow2014,Cai2021} based on reconstruction \cite{Li2018,Li2019} and RNN-based forecasting approaches \cite{Hundman2018,Zhao2020} have also reported promising performance for multivariate anomaly detection. However, \emph{these methods do not explicitly learn the topological structure among sensors, thus leaving room for improvements in modeling high-dimensional sensor data with considerable potential inter-relationships appropriately}. 

Graph Convolutional Networks (GCNs) \cite{Zhou2018,Velickovic2018,Wu2019,Wang2019} have recently revealed discriminative power in learning graph representations due to their permutation-invariance, local connectivity, and compositionality \cite{Zhou2018,Wu2020}. Graph neural networks allow each graph node to acknowledge its neighborhood context by propagating information through structures. Recent works \cite{Zhao2020,Cao2020,Deng2021} then combined temporal modeling methods with GCNs to model the topological relationships between sensors. Specifically, most existing graph-based approaches \cite{Wu2020,Deng2021} aim to learn the graph structure by measuring the cosine similarity (or other distance metrics) between sensor embeddings and defining top-K closest nodes as the source node's \emph{connections}, followed by a graph attention convolution to capture the information propagation process. However, we argue that (1) \emph{dot products among sensor embeddings lead inevitably to quadratic time and space complexity regarding the number of sensors}; (2) \emph{the tightness of spatial distance can not entirely indicate that there exists a strong connection in a topological structure}. 

To address the problems above, we propose an innovative framework named \emph{\textbf{G}raph Learning with \textbf{T}ransformer for \textbf{A}nomaly detection} (GTA) in this paper. We devise from the perspective of learning a global bi-directed graph structure involving all sensors within the entity through a \emph{connection learning policy} based on the Gumbel-Softmax Sampling trick to overcome the quadratic complexity challenge and the limitations of top-K \emph{nearest} strategy. The policy logits can automatically discover the hidden associations during the training process by determining whether any specific node's information should flow to the other targets to achieve the best forecasting accuracy while restricting each node's neighborhoods' scope as much as possible. The discovered hidden associations are then fed into the graph convolution layers for information propagation modeling. We then integrate these graph convolution layers with different level dilated convolution layers to construct a hierarchical context encoding block specifically for temporal data. \emph{While recurrent mechanisms can be naturally applied to temporal dependency modeling, it is hard to parallelize in many mobile environments (e.g., IoT), which require high computation efficiency.} Hence we adopt the \emph{Transformer} \cite{Vaswani2017} based architecture for the sequence modeling and forecasting due to the parallel efficiency and capability of capturing long-distance context information. We also propose a novel multi-branch attention strategy to reduce the quadratic complexity of original self multi-head attention. 

The main \textbf{contributions} of our work are summarized as follows:
\begin{itemize}
    \item We propose a novel and differentiable \emph{connection learning policy} to automatically learn the graph structure of dependency relationships between sensors. Meanwhile, each node's neighborhood field is restricted by integrating a new loss term for further inference efficiency.
   \item We introduce a novel graph convolution named Information Propagation (IP) convolution to model the anomaly influence flowing process. A multi-scale dilated convolution is then combined with the graph convolution to form an effective hierarchical temporal context encoding block.
   \item We propose a novel multi-branch attention mechanism to tackle the original multi-head attention mechanism's quadratic complexity challenge. 
   \item We conduct extensive experiments on a wide range of multivariate time series anomaly detection benchmarks to demonstrate the superiority of our proposed approach over state-of-the-arts. 
\end{itemize}

\section{Related Work}
The existing literature for 
% There are a great deal of literature for 
addressing time series anomaly detection usually can be  divided into two major categories. 
% considering each variable as independent of the others
The first category usually modeled each time series variable independently, while 
% each variable as independent of the others and modeled each time series for modeling. From a more intuitive perspective, 
the second category took into consideration the correlations among multivariate time series to improve the performance. 
% rather than isolating multiple time series. 
% Among the aforementioned categories, the algorithm of mining the relationship between variables based on Graph Neural Network (GNN)[1] is enlightening to our work. Thus, additionally, we will elaborate those state-of-art approaches that complete anomaly detection based on GNN in the last section. 

\subsection{Anomaly Detection in Univariate Time Series}
The anomaly detection in univariate time series has drawn many researchers' attentions in recent years. 
Traditionally, 
% The majority of classical univariate 
the anomaly detection frameworks included two main phases: estimation phase and detection phase \cite{garcia2020}. In estimation phase, the variable values at one timestamp or time interval can be  predicted or estimated by specific algorithm. 
Then the estimated values were compared with real  
values based on dynamically adjusted thresholds 
% which is preseted or , 
to detect anomalies in detection phase. 
For example, Zhang et. al \cite{Zhang2005} applied  ARIMA  
% to their framework 
to capture the linear dependencies between the future values and the past values, thus modeling the time series behavior for anomaly detection. 
Lu et.al \cite{Lu2009} utilized wavelet analysis to construct the estimation model.  
% and measure the difference. 
With the development of deep learning, various neural network architectures have also been applied to anomaly detection. 
DeepAnt \cite{Munir2019} was an unsupervised approach using  convolutional neural network (CNN) to forecast future time series values and adopted Euclidean distance to measure the discrepancy for anomaly detection. 
The LSTM neural network was also widely used in modeling time series behaviors  \cite{Malhotra2016,Filonov2016,Li2019}.  
The LSTM-based encoder-decoder \cite{Malhotra2016}  reconstructed the variable values and measured the  reconstruction errors for detection. 
% Filonov et. al \cite{Filonov2016} adopted LSTM to their framework, which shows a good performance on forecasting, to predict the expected values. 
% Malhotra et.al \cite{Malhotra2015} used stacked LSTM networks to model the normal behavior which captured  the time sequence regularity and averts the  pre-determined time window. 
\begin{figure*}[htb]
    \centering
    \includegraphics[width=.75\linewidth]{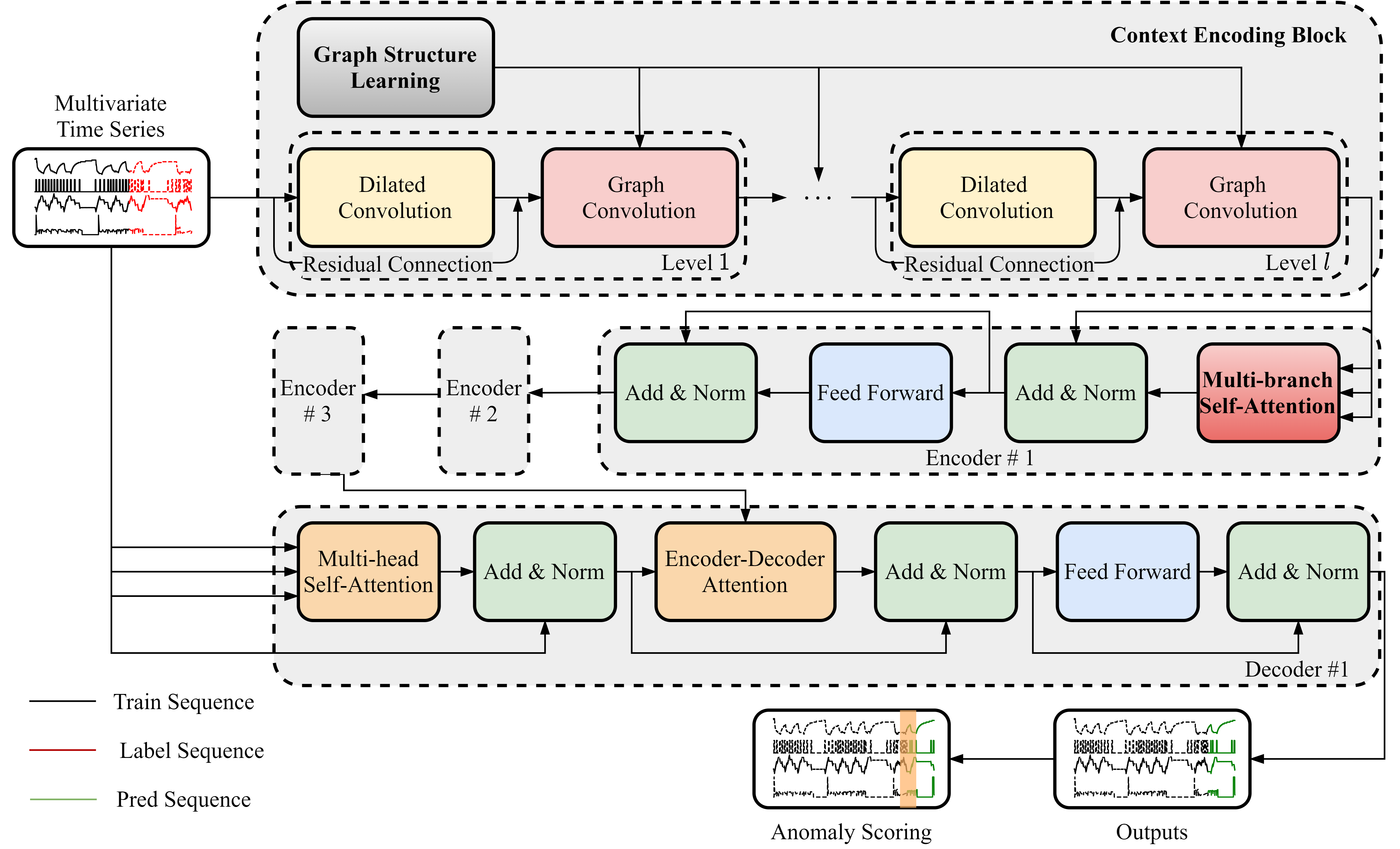}
    \caption{The visualization of our proposed GTA's architecture with $l$ levels dilated convolution and graph convolution, 3 encoder layers, and 1 decoder layer. Generally, the input multivariate time series inputs are split into train sequences and label sequences, of which train sequences are fed into encoder while label sequences are fed to the decoder.}
    \label{fig:model}
\end{figure*}

\subsection{Anomaly Detection in Multivariate Time Series}
% In the actual deployment of the IoT networks, it is unrealistic for a single data acquisition source to realize environmental monitoring[6]. 
In real-world scenarios, the time series data acquisition sources could be multiple \cite{Cook2020}. 
Therefore, many work began to pay attention to exploiting  the correlations among multiple variables  to improve the accuracy of anomaly detection. 
Jones et.al \cite{Jones2016} extracted statistical and smoothed trajectory (SST) features of time series and utilized a set of non-linear functions to model related variables to detect anomalies. 
% to representante  
% a neoteric representation - - to find anomaly and use a set of non-linear function to model related variables. 
Using the LSTM network as the base models to 
to capture the temporal correlations of time series data,  MAD-GAN \cite{Li2019} proposed an unsupervised anomaly detection method 
combining generative adversarial networks (GAN) by 
considering complex  dependencies amongst different time series variables. 
Sakurada et al. \cite{Sakurada2014} conducted  dimentionality reduction based on autoencoders for anomaly  detection. 
 The ODCA framework \cite{Lu2018} included three parts: data preprocessing, outlier analysis, and outlier rank, which  used cross correlation to translate  high-dimentional data sets to one-dimentional  cross-correlation function. 
%  both of which weaken the correlation between variables to a certain extent due to dimentional reduction. 
OmniAnomaly \cite{Su2019} was a stochastic model to avoid potential misguiding by uncertain instances, which used  stochastic variable connection and normalizing flow to get reconstruction probabilities and adopted streaming POT with drift (DSPOT) algorithm \cite{Siffer2017} for automatic threshold selection. Senin et al. \cite{Senin2015} proposed two algorithms that conducted symbolic time series discretization and used grammar reduction to compress the input sequence and compactly encode them with grammar rules. Those rarely used substrings in the grammar rules were regarded as anomalies. Autoregressive with exogenous inputs (ARX) and artificial neural network (ANN) \cite{Akouemo2017} extracted time-series features and detected anomalous data points by conducting hypothesis testing on the extrema of residuals.

% \subsection{Graph Neural Network(GNN) based Anomaly Detection}
% Aforementioned approaches somehow ignore or not sufficiently explore the correlation between multivariables. 
% In recent years, the appearance of graph neural network  (GNN) showed good performance in modeling graph structure, which provided a new way of anomaly detection. 
To cover the shortage that the convolution and pooling operators of CNNs are defined for regular grids, recent GNN \cite{Defferrard2016} generalizes CNNs to graphs that are able to encode irregular and non-Euclidean structures. GNN adopted the localized spectral filters and used a graph coarsening algorithm to cluster similar vertices for speeding up. In this way, GNN efficiently extracted the local stationary property and captured the correlation between nodes.
In real-world IoT environment, the graph structure modeling the correlations between sensors is often not predefined in advance. 
Graph deviation network (GDN) \cite{Deng2021} learned  the pairwise relationship by cosine similarity to  elaborate adjacent matrix which can be modeled as a graph. 
% and translates it into edge representation. 
Then it predicted the future values by graph attention-based forecasting and computed the absolute error value to evaluate graph deviation score. 
MTAD-GAN \cite{Zhao2020} concatenated feature-oriented and time-oriented graph attention layer to learn graph structure and used both forecasting-based model and  reconstruction-based model to calculate integrated loss.   
% With inference score evaluated from integrated loss, 
Then automatic threshold algorithm was adopted  to  perform anomaly detection.

\section{Problem Statement}
In this work, we focus on the task of multivariate time series anomaly detection. Let $\mathcal{X}^{(t)}\in \mathbb{R}^{M}$ denote the original multivariate time series data at any timestamp $t$, where $M$ is the total number of sensors or any data measuring node within the same entity. $M$ is also reported as the number of features or variables in some literature \cite{Li2019,Su2019,Deng2021}. Considering the high unbalance between normal data and anomalies, we only construct the sequence modeling process on normal data (without anomalies) and make prediction on testing data (with anomalies) for anomaly detection. Specifically, we let $\mathcal{X}$ and $\hat{\mathcal{X}}$ represent the entire normal data and data with anomalies, respectively. For sequence modeling on normal data, we inherit a forecasting-based strategy to predict the time series value $\mathbf{x}^{(t)}\in \mathbb{R}^{M}$ at next time step $t$ (aka. single-step time series forecasting) based on the historical data $\mathbf{x}=\{\mathbf{x}^{(t-n)}, \cdots, \mathbf{x}^{(t-1)}\}$ with a specific window size $n$. Therefore, given a sequence of historical $n$ time steps of multivariate contiguous observations $\hat{\mathbf{x}}\in \mathbb{R}^{M\times n}$, the goal of anomaly detection is to predict the output vector $\hat{\mathbf{y}}\in \mathbb{R}^{n}$, where $\hat{\mathbf{y}}^{(t)}\in \{0, 1\}$ denotes binary labels indicating whether there is an anomaly at time tick $t$. Precisely, our proposed approach returns an anomaly score for each testing timestamp, and then the anomaly result can be obtained via selecting different thresholds. 

We also provide some basic graph-related concepts for better understanding formulated as follows:
\begin{definition}[Graph]
A directed graph is formulated as $\mathcal{G}=(\mathcal{V}, \mathcal{E})$ where $\mathcal{V}=\{1, \cdots, M\}$ is the set of nodes, and $\mathcal{E}\subseteq \mathcal{V}\times \mathcal{V}$ is the set of edges, where $\mathbf{e}_{i, j}$ represents the uni-directed edge flowing from node $i$ to node $j$.
\end{definition}
\begin{definition}[Node Neighborhood]
Let $i\in \mathcal{V}$ denote a node and $\mathbf{e}_{i, j}\in \mathcal{E}$ denote the edge pointing from node $i$ to node $j$. The neighborhood of any node $i$ is defined as $\mathcal{N}(i)=\{j\in \mathcal{V}\vert \mathbf{e}_{i, j}\in \mathcal{E}\}$.
\end{definition}
% \begin{definition}[Graph]
% A directed graph is formulated as $\mathcal{G}=(\mathcal{V}, \mathcal{E})$ where $\mathcal{V}=\{1, \cdots, N\}$ is the set of nodes, and $\mathcal{E}\subseteq \mathcal{V}\times \mathcal{V}$ is the set of edges, where $\mathbf{e}_{i, j}$ represents the uni-directed edge connection from node $i$ to node $j$.
% \end{definition}

\section{Methodology}
In most real-life scenarios of IoT, there are usually complex topological relationships between sensors where the entire entity can be seen as a graph structure. Each sensor is also viewed as a specific node in the graph. Previous methods \cite{Wu2020,Deng2021} focused on applying various distance metrics to measure the relations between nodes mostly by selecting the top-K closest ones as their neighbor dependencies. Different from existing approaches, we devise a \emph{directed graph structure learning policy} (see Fig. \ref{fig:model}) to automatically learn the adjacency matrix among nodes such that the network can achieve the maximum benefits. The core of the learning policy is named \emph{Gumbel-Softmax Sampling} strategy \cite{Maddison2014,Jang2017} inspired by the policy learning network in many reinforcement learning methods \cite{Rosenbaum2018,Guo2019}. These discovered hidden associations are then fed into the graph convolution layers for information propagation modeling. We then integrate these graph convolution layers with different level dilated convolution layers together to construct a \emph{hierarchical context encoding block} specifically for temporal data. The outputs of the context encoding block are then applied positional encoding as the inputs of \emph{Transformer} \cite{Vaswani2017} for single-step time series forecasting. We also propose a global attention strategy to overcome the quadratic computation complexity challenge of the multi-head attention mechanism. Fig. \ref{fig:model} further illustrates the entire architecture in detail.

\subsection{Gumbel-Softmax Sampling}
\begin{figure}
    \centering
    \includegraphics[width=.4\linewidth]{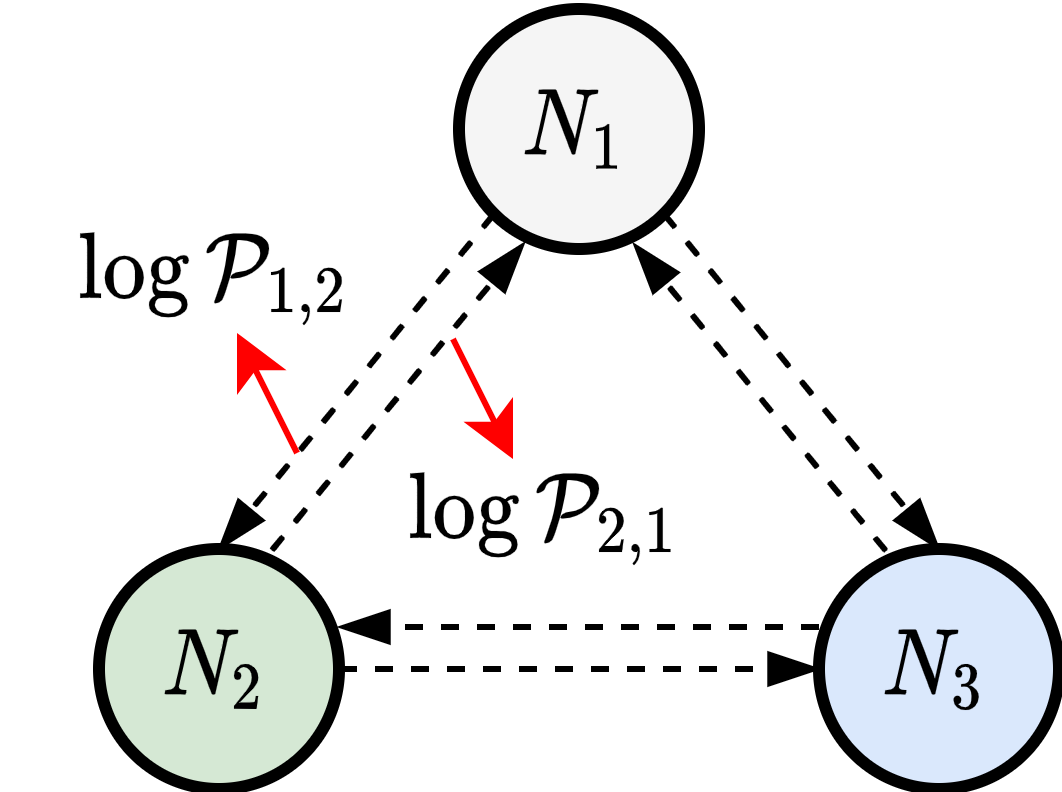}
    \caption{Suppose we have 3 sensors ($N_{1}, N_{2}, N_{3}$) of which the dependencies are yet hidden. Our connection learning policy's main idea is to use the Gumbel-Softmax Sampling strategy to sample a random categorical vector for determining whether any directed connection between two nodes can be established. For $N_{1}$ and $N_{2}$, if the value of $\mathcal{P}_{1,2}$ is relatively high, it represents $N_{1}$ is highly possibly pointed to $N_{2}$, vice versa.}
    \label{fig:logits}
\end{figure}
The sampling process of discrete data from a categorical distribution is originally non-differentiable, where typical backpropagation in deep neural networks cannot be conducted. \cite{Maddison2014,Jang2017} proposed a differentiable substitution of discrete random variables in stochastic computations by introducing Gumbel-Softmax distribution, a continuous distribution over the simplex that can approximate samples from a categorical distribution. In our graph learning policy with a total number of $M$ candidate nodes, we let $\mathbf{z}^{i, j}$ be a binary connection control variable for any pair of nodes $i$ and $j$ with uni-directed probabilities from node $i$ to node $j$ as $\{\pi^{i, j}_{0}, \pi^{i, j}_{1}\}$, where $\pi^{i, j}_{0} + \pi^{i, j}_{1} = 1$ and $\pi^{i, j}_{1}$ represents the probability that there exists an information flow from node $i$ to node $j$ in the graph (see Fig. \ref{fig:logits}). Similarly, by Gumbel-Max trick, we can sample any pair of nodes' connection strategy $z^{i, j}\in \{0, 1\}^{2}$ with:
\begin{equation}
    z^{i, j} = \argmax_{c\in \{0, 1\}}(\log \pi^{i, j}_{c} + g^{i, j}_{c})
\end{equation}
where $g_{0}, g_{1}$ are i.i.d samples drawn from a standard Gumbel distribution which can be easily sampled using inverse transform sampling by drawing $u \sim$ Uniform$(0, 1)$ and computing $g=-\log(-\log u)$. We further substitute this $\argmax$ operation, since it is not differentiable, with a $\mathrm{Softmax}$ reparameterization trick, also known as Gumbel-Softmax trick, as:
\begin{equation}
    z^{i, j}_{c} = \frac{\exp((\log \pi^{i, j}_{c} + g^{i, j}_{c})/\tau)}{\sum\limits_{v\in \{0, 1\}}\exp((\log \pi^{i, j}_{v} + g^{i, j}_{v})/\tau)}
\end{equation}
where $c\in \{0, 1\}$ and $\tau$ is the temperature parameter to control Gumbel-Softmax distribution's smoothness, as the temperature $\tau$ approaches 0, the Gumbel-Softmax distribution becomes identical to the one-hot categorical distribution. As the randomness of $g$ is independent of $\pi$, we can now directly optimize our gating control policy using standard gradient descent algorithms. 

Compared to the previous graph structure learning approaches, our proposed method significantly reduces the computation complexity from $\mathcal{O}(M^2)$ to $\mathcal{O}(1)$ since it requires no dot products among high-dimensional node embeddings. Additionally, the graph structure learning policy is able to automatically learn the global topological connections among all nodes, thereby avoiding the limitation of selecting only the top-K \textit{nearest} nodes as neighbors.

\subsection{Influence Propagation via Graph Convolution}
On top of the learned topological structure, the graph convolution block aims to further model the influence propagation process and update each specific node's representation by incorporating its neighbors' information. Considering the characteristics of tasks such as anomaly detection, usually, the occurrence of abnormalities is due to a series of chain influences caused by one or several nodes being attacked. Therefore, it is intuitive for us to model the relationships between upstream and downstream nodes by capturing both temporal and spatial differences. Thus, we define our Influence Propagation (IP) convolution process concerning each specific node and its neighborhoods by applying a node-wise symmetric aggregating operation $\square$ (e.g., add, mean, or max) on the differences between nodes associated with all the edges emanating from each node. The updated output of IPConv at the $i$-th node is given by:
\begin{equation}
    \mathbf{x}^{\prime}_{i} = \sum\limits_{j\in \mathcal{N}(i)}h_{\mathbf{\Theta}}(\mathbf{x}_{i}\vert\vert \mathbf{x}_{j}-\mathbf{x}_{j}\vert\vert \mathbf{x}_{j}+\mathbf{x}_{i})
\end{equation}
where $\square$ is chosen as summation in our method, $h_{\mathbf{\Theta}}$ denotes a neural network, i.e. MLPs (Multi-layer Perceptrons), $\mathbf{x}_{i}\in \mathbb{R}^{T}$ represents the time series embedding of node $i$ and $\vert\vert$ denotes the concatenation operation. We denote $\mathbf{x}_{j}-\mathbf{x}_{i}$ as the differences between nodes to \emph{explicitly model the influence propagation delay} from node $j$ to $i$, captured by the value difference at each timestamp of the time series embedding. We also incorporate the term $\mathbf{x}_{i}+\mathbf{x}_{j}$ with the differences to work as a \emph{scale benchmark} such that the model can learn the truly generalized impact to the other nodes brought by anomalies instead of extreme values. Intuitively, for any specific node $i$, if one of its neighbor nodes $j$ being attacked, node $i$ shall be severely affected sooner or later due to the restricted topological relationship. 

\textbf{Training Strategy and Regularization.} Graph convolution based on the learned dependencies among sensors only aggregates the information from nodes' neighbors without taking efficiency into account. Under the extreme circumstance that all nodes are mutually connected, aggregating neighborhood information adds considerable noise to each node. However, it is preferred to form a compact sub-graph structure for every single node, in which redundant connections are omitted as much as possible without deteriorating the forecasting accuracy. To this end, we propose a sparsity regularization $\mathcal{L}_{s}$ to enhance the compactness of each node by minimizing the log-likelihood of the probability of a connection being established as
\begin{equation}
    \mathcal{L}_{s} = \sum\limits_{1\leq i,j \leq M, i\neq j} \log \pi^{i, j}_{1}
\end{equation}
Furthermore, to encourage better learning convergence, the \emph{connection learning policy} is initialized with all nodes connected. We \emph{warm up} the network weights by training with this \emph{complete graph structure} for a few epochs to provide a good starting point for the policy learning. 
\begin{figure}
    \centering
    \includegraphics[width=.6\linewidth]{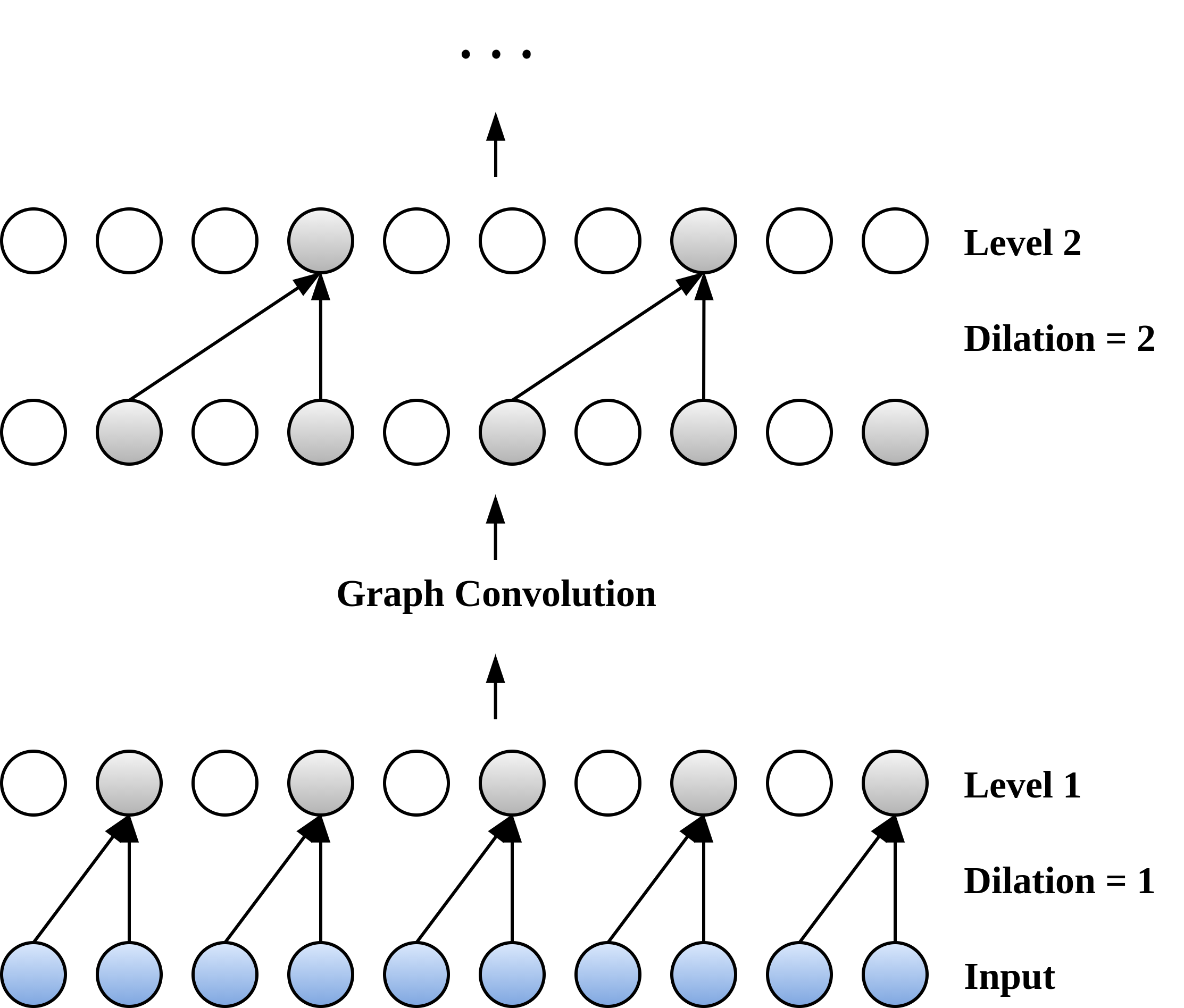}
    \caption{Visualization of hierarchical dilated convolution combined with graph convolution.}
    \label{fig:dilated}
\end{figure}

\begin{figure*}[htb]
%   \vspace{-2.5cm}
\centering
  \begin{subfigure}[b]{.5\columnwidth}
    % \hspace{-1.5cm}
    \includegraphics[width=.8\linewidth]{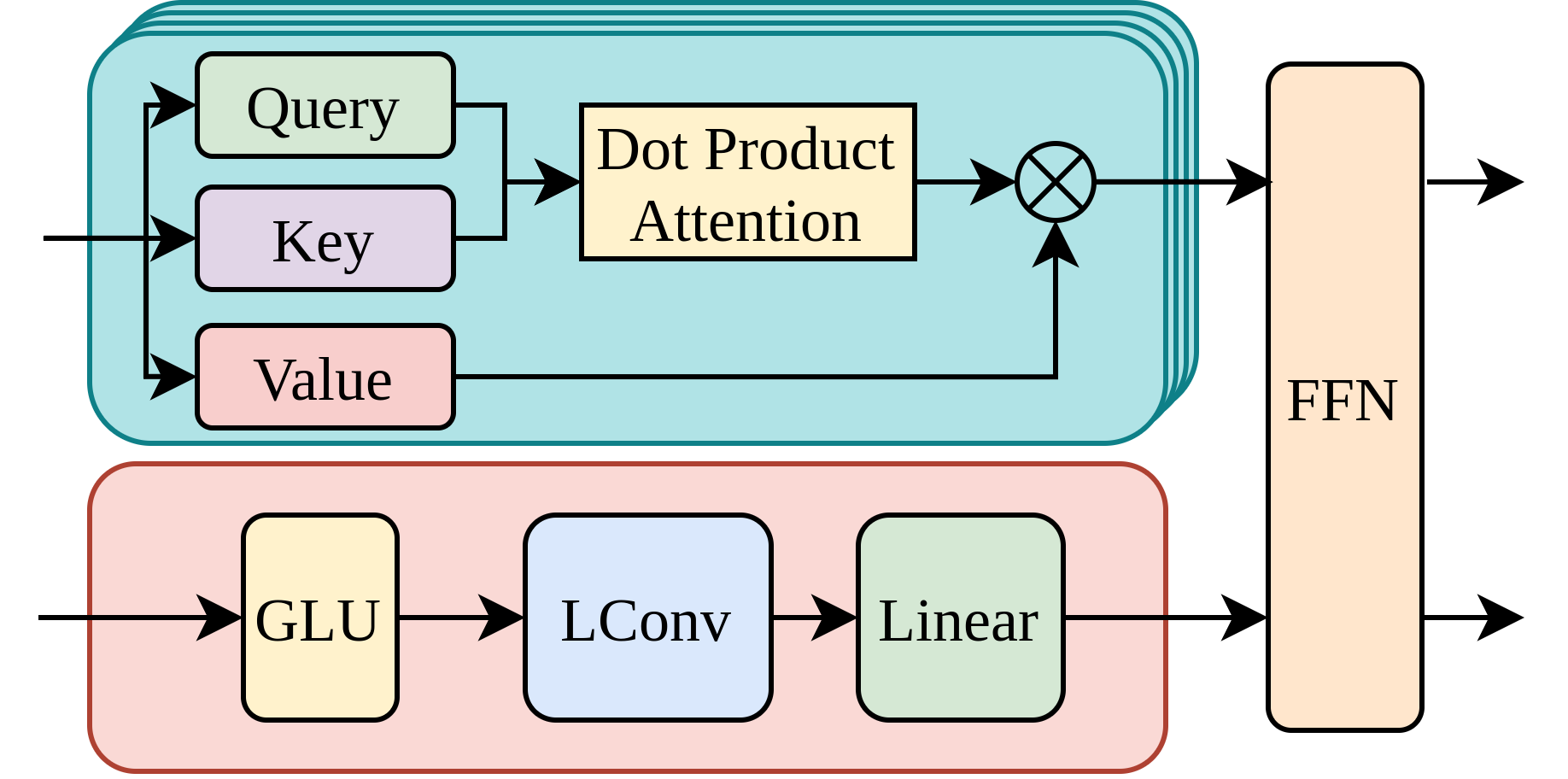}
    % \hspace{1.5cm}
    % \vspace{-4cm}
    \caption{Vanilla multi-branch Transformer.}
    \label{fig:t1}
  \end{subfigure}
  \begin{subfigure}[b]{.5\columnwidth}
    % \hspace{-1cm}
    % \hspace{0.1\textwidth}
    \includegraphics[width=.8\linewidth]{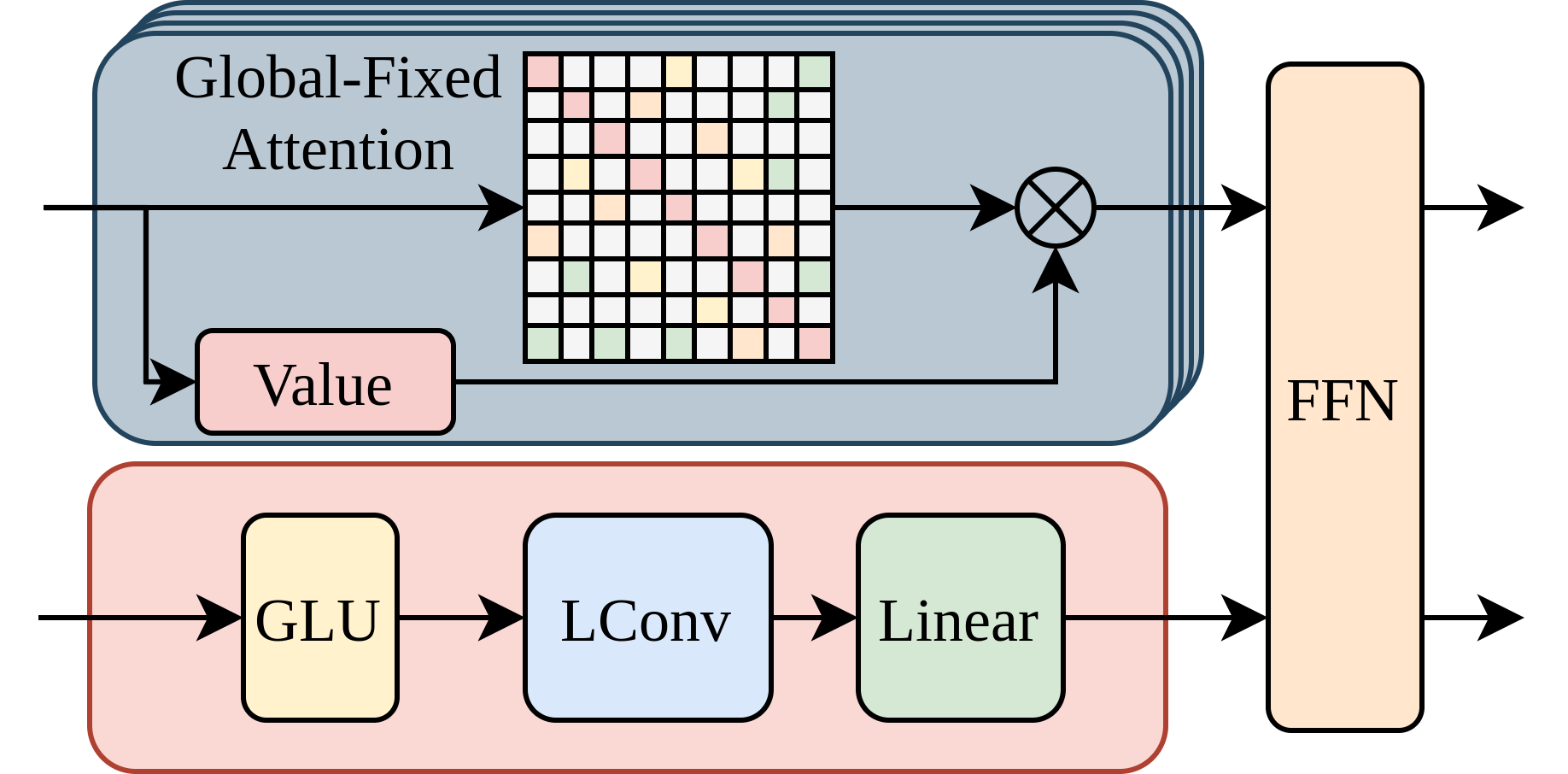}
    \caption{Multi-branch Transformer with global-fixed attention.}
    \label{fig:t2}
  \end{subfigure}
%   \hspace{0.1\textwidth}
%   \hfill
  \begin{subfigure}[b]{.5\columnwidth}
    % \hspace{-1cm}
    \includegraphics[width=.8\linewidth]{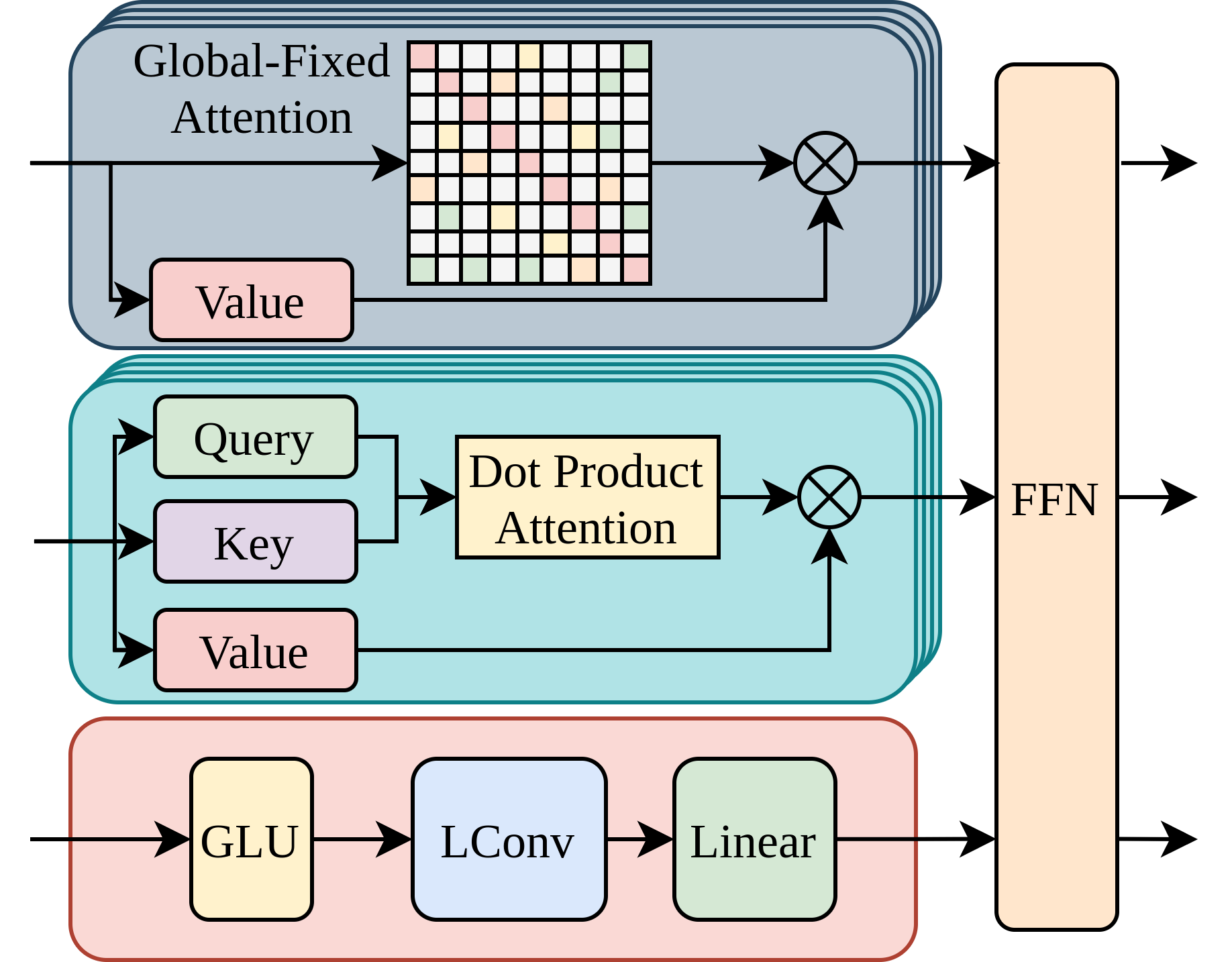}
    \caption{Our proposed multi-branch attention mechanism.}
    \label{fig:t3}
  \end{subfigure}
  \caption{Different variants of efficient multi-branch attention mechanism. Right: Replacing vanilla multi-head attention with a combination of both global-fixed attention and vanilla multi-head attention and neighborhood convolution by splitting embeddings into multiple channels.} 
  \label{fig:attention}
\end{figure*}

\subsection{Hierarchical Dilated Convolution}
The dilated convolution \cite{Yu2016} is widely used in sequence modeling due to its powerful capability in extracting high-level temporal context features by capturing sequential patterns of time series data through standard 1D convolution filters. Setting different dilation size levels can discover temporal patterns with various ranges and handle very long sequences. However, choosing the right kernel size is often a challenging problem for convolutional operations. Some previous approaches adopted the widely employed inception learning strategy \cite{Szegedy2015} in computer vision which concatenates the outputs of convolutional filters with different kernel sizes followed by a weighted matrix. Unlike them, we propose a \emph{hierarchical dilated convolution learning strategy} combined with the graph convolution above to fully explore the temporal context modeling process with different sequence lengths and receptive fields by setting multi-scale dilation sizes. Specifically, as Fig. \ref{fig:dilated} illustrates, the bottom layer represents the multivariate time series input (for some time $t$, onto which repeated dilated convolutions, with increasing dilation rates, are applied; the filter width is again set to equal two in the observed model). The first level block applies dilated convolutions with the dilation rate equal to one, meaning that the layer applies the filter onto two adjacent elements, $\mathbf{x}^{(t)}$ and $\mathbf{x}^{(t+1)}$, of the input series. The outputs of the first-level dilated convolutions are fed into the graph convolution module proposed above. Then the second-level layer applies dilated convolutions, with the rate now set to equal two, which means that the filter is applied onto elements $\mathbf{x}^{(t)}$ and $\mathbf{x}^{(t+2)}$ (notice here that the number of parameters remains the same, but the filter width has been “widened”). By setting multi-scale dilation sizes with a hierarchical learning style, abundant temporal representations concerning different temporal positions and sequence lengths can be effectively learned. 

The hierarchical dilated convolution and the graph convolution together form the temporal context embedding progression where dilated convolution captures the long-term temporal dependencies while graph convolution describes the topological connection relationships between sensors (or nodes). As a result, the final outputs have been well represented to be the inputs of the next forecasting procedure using Transformer architecture. 

\subsection{More Efficient Multi-branch Transformer}
Transformer \cite{Vaswani2017} has been widely used in sequence modeling due to the superior capability of multi-head attention mechanism in long-distance dependencies capturing. However, one main efficiency bottleneck in self-attention is that the pairwise token interaction dot-production incurs a complexity of $\mathcal{O}(n^2)$ with respect to sequence length. To tackle this challenge, in this section, we first briefly review the background of some recent development of multi-head attention mechanism and then propose a more efficient Transformer architecture based on the innovative multi-branch attention mechanism which is more computationally efficient. 

\textbf{Self-attention in Transformers.}
The vanilla multi-head self-attention mechanism was originally proposed by \cite{Vaswani2017}. For a sequence of token representations $\mathbf{X}\in \mathcal{R}^{n\times d}$ (with sequence length $n$ and dimensionality $d$), the self-attention function firstly projects them into queries $\mathbf{Q}\in \mathcal{R}^{n\times d_{k}}$, keys $\mathbf{K}\in \mathcal{R}^{n\times d_{k}}$ and values $\mathbf{V}\in \mathcal{R}^{n\times d_{v}}$, $h$ times with different, learned linear projections to $d_{k}$, $d_{k}$ and $d_{v}$ dimensions, respectively. Then a particular scaled dot-product attention was computed to obtain the weights on the values as:
\begin{equation}
    \mathrm{Attention}(\mathbf{Q}, \mathbf{K}, \mathbf{V})=\mathrm{Softmax}(\frac{\mathbf{Q}\mathbf{K}^{T}}{\sqrt{d_{k}}})\mathbf{V}
\end{equation}
Multi-head attention allows the model to jointly attend to information from different representation subspaces at different positions. With a concatenated computing way, the final output of multi-head attention is as following:
\begin{equation}
    \mathrm{MultiHead}(\mathbf{Q}, \mathbf{K}, \mathbf{V})=\mathrm{Concat}(\textit{head}_{1}, \cdots, \textit{head}_{h})W^{O}
\end{equation}
in which, $h$ is the number of total heads.
Each head is defined as:
\begin{equation}
    \textit{head}_{i}=\mathrm{Attention}(\mathbf{Q}W_{i}^{Q}, \mathbf{K}W_{i}^{K}, \mathbf{V}W_{i}^{V})
\end{equation}
where the projections are parameter matrices $W_{i}^{Q}\in \mathcal{R}^{d\times d_{k}}$, $W_{i}^{K}\in \mathcal{R}^{d\times d_{k}}$, $W_{i}^{V}\in \mathcal{R}^{d\times d_{v}}$ and $W^{O}\in \mathcal{R}^{hd_{v}\times d}$. 

\textbf{Global-learned Attention.}
Recent research \cite{Raganato2018,Voita2019} claim that the self-attention in Transformers can be substantially simplified with trivial attentive patterns at training time: only preserving adjacent and previous tokens is necessary. The adjacent positional information such as "current token", "previous token" and "next token" are the key features learned across all layers by encoder self-attention. Instead of costly learning the trivial pattern using massive corpus with considerable computational resources, the conventional pairwise token interaction attention could be replaced by a more computation-efficient global attention pattern. In practice, manually pre-define all global-fixed patterns is easy to implement but can barely cover all possible situations. To generalize the global-fixed attention pattern proposed in \cite{Raganato2020}, we apply a parameter matrix $\mathbf{S}\in \mathcal{R}^{m\times m}$ ($m>n$) as a learnable global compatibility function across all training samples following the \emph{Synthesizer} \cite{Tay2020}. Hence, each head adds $m^2$ parameters while reducing two projection matrices $W^{Q}$ and $W^{K}$. The attention now has been as following:
\begin{equation}
    \mathrm{Attention}(\mathbf{S}, \mathbf{V})=\mathrm{Softmax}(\mathbf{S})\mathbf{V}
\end{equation}
where $\mathbf{S}$ is a learnable matrix which can be randomly initialized.

\begin{table*}[htb]
    \centering
    \caption{Memory and computation analysis on different attention types.}
    \begin{tabular}{@{}lccc@{}}
    \toprule
    Attention Type     & $|\theta|$ &  \# Mult-Adds & Global/Inter \\ \midrule
    Scaled Dot-Product & $4d^{2}$   &  $\mathcal{O}(4nd^{2}+2n^{2}d)$ & Inter \\
    Global-Learned     & $m^{2}h+2d^{2}$ & $\mathcal{O}(2nd^{2}+n^{2}d)$ & Global \\
    % Head-Wise Mixing              & $m^2h_{1}+2dd_{k}h_{2}+2d^{2}$ & $\mathcal{O}(2nd^{2}+n^{2}d+2ndd_{k}h_{1}+n^{2}d_{k}h_{1})$ & Both \\
    Branch-Wise Mixing     & $4d_{1}^{2}+m^{2}h+2d_{2}^{2}$ &
    $\mathcal{O}(4nd_{1}^{2}+n^2d_{1}+2nd_{2}^{2}+n^2d)$ & Both \\
    \bottomrule
    \end{tabular}
    \label{tab:attention}
\end{table*}

\textbf{Multi-branch Architecture for Transformers.}
\cite{Zhanghao2020} has demonstrated the effectiveness of multi-branch attention in capturing global and local context patterns, especially under mobile computational constraints. As Fig. \ref{fig:attention} illustrates, this double-branch architecture splits the original input sequences into two pieces along the embedding channel, followed by two attention branches: one convolution branch for extracting information in a restricted neighborhood and one multi-head attention branch for capturing long-distance dependencies. As a substitute for vanilla self-attention, we apply a task-specific alignment matrix that learns globally across all training samples where attention weights are no longer conditioned on any input token in our architecture. By simply replacing the dot-production with global-learned alignment as Fig. \ref{fig:attention} shows, the input sequences will only be projected into \emph{value} matrices. A weighted sum up of \emph{values} is then calculated using this global-learned attention. In order to explore a better trade-off between computation efficiency and model performance, we propose to combine the pairwise token interactions and global-learned attention in terms of a branch-wise mixing strategy.

\textbf{Branch-wise Mixing.} For branch-wise mixing, the input sequences are split into multiple branches along the embedding dimension as Fig. \ref{fig:attention} clearly describes. Different from the original two-branch architecture, we build one more branch for global-learned attention. Thus,
\begin{equation}
    \begin{split}
        \mathrm{Attention}&=\mathrm{Concat}(\mathbf{A}^{(1)}, \mathbf{A}^{(2)}) \\
        \mathbf{A}^{(1)}&=\mathrm{MultiHead}(\mathbf{X}^{(1)}) \\
 \mathbf{A}^{(2)}&=\mathrm{Global}(\mathbf{X}^{(2)})      
    \end{split}
\end{equation}
where $\mathbf{X}^{(1)}\in \mathcal{R}^{n\times d_{1}}$, $\mathbf{X}^{(2)}\in \mathcal{R}^{n\times d_{2}}$ and $d=d_{1}+d_{2}$.

In our models, we only change the branch that captures the global contexts while remaining the local pattern extractor using either \textit{lightConv} or \textit{dynamicConv} \cite{Felix2019}.

\textbf{Computation Analysis.}
Table \ref{tab:attention} lists the different model variants explored within our proposed framework. The column $|\theta|$ refers to the total number of parameters in one self-attention module excluding the feed-forward layer. Obviously, compared to the original scaled dot-production, the amount of computation of global-learned attention is directly reduced by half in terms of Mult-Adds. Our proposed multi-branch mixing strategy increases the amount of calculation in varying degrees due to the mix with scaled dot-production. However, this is a trade-off between computation complexity and model size. More precisely, when $m \leq \sqrt{2/h}d$, the global attention module is more computationally efficient than the other variants.

\subsection{Anomaly Scoring}
Inspired by \cite{Zhou2020}, the original multivariate time series inputs are split into two parts: training sequences for the encoder and label sequences for the decoder. The decoder receives long sequence inputs, pads the target elements into zero, measures the weighted attention composition of the feature map, and instantly predicts output elements in a generative style. Let the single-step prediction denote as $\hat{\mathcal{Y}}\in \mathbb{R}^{M\times n}$. We apply the Mean Square Error (MSE) between the predicted outputs $\hat{\mathbf{Y}}$ and the observation $\mathcal{Y}$, as the loss function to minimize:
\begin{equation}
    \mathcal{L}_{mse} = \frac{1}{M}\sum\limits_{t=1}^{n}\vert\vert\mathcal{Y}^{(t)} - \hat{\mathcal{Y}}^{(t)}\vert\vert^{2}_{2}
\end{equation}

Similar to the loss objective, the anomalous score compares the expected value at time $t$ to the observed value, computing an anomaly score via the deviation level as:
\begin{equation}
    \hat{\mathbf{y}}^{(t)} = \sum\limits_{i=1}^{M}\vert\vert\mathcal{Y}_{i}^{(t)} - \hat{\mathcal{Y}}_{i}^{(t)}\vert\vert^{2}_{2}
\end{equation}
Finally, we label a timestamp $t$ as an anomaly if $\hat{\mathbf{y}}^{(t)}$ exceeds a fixed threshold. Since different approaches could be employed to set the threshold such as extreme value theory \cite{Siffer2017}, the same anomaly detection model could result in different prediction performance with different anomaly thresholds. Thus, we apply a grid search on all possible anomaly thresholds to search for the best $\mathrm{F1}$-score (with notation $^{**}$) and $\mathrm{Recall}$ (with notation $^{*}$) in theory and report them.

\section{Experiments}
\subsection{Datasets}
We evaluate our method over a wide range of real-world anomaly detection datasets. \textbf{SWaT} \cite{Mathur2016} The Secure Water Treatment dataset is collected from a water treatment testbed for cyber-attack investigation initially launched in May 2015. The SWaT dataset collection process lasted for 11 days, with the system operated 24 hours per day such that the network traffic and all the values obtained from all 51 sensors and actuators are recorded. Due to the system working flow characteristics, there is a natural topological structure relationship between all sensing nodes. After this, a total of 41 attacks derived through an attack model considering the intent space of a CPS were launched during the last 4 days of the 2016 SWaT data collection process. As such, the overall sequential data is labeled according to normal and abnormal behaviors at each timestamp. \textbf{WADI} \cite{Ahmed2017} Water Distribution dataset is collected from a water distribution testbed as an extension of the SWaT testbed. It consists of a total of 16 days of continuous operations with 14 days under regular operation and 2 days with attack scenarios. The entire testbed contains 123 sensors and actuators. Moreover, \textbf{SMAP} (Soil Moisture Active Passive satellite) and \textbf{MSL} (Mars Science Laboratory rover) are two public datasets published by NASA \cite{Oneill2010}. Each dataset has a training and a testing subset, and anomalies in both testing subsets have been labeled \cite{Hundman2018}. 

Table \ref{tab:wadi} and \ref{tab:nasa} summarises the statistics of the four datasets. In order to fair comparison with other methods, the original data samples for \textbf{SWaT} and \textbf{WADI} are downsampled to one measurement every 10 seconds by taking the median values following \cite{Deng2021}.

%----------------------------------------------------------------------------------
\begin{table}[htb]
\caption{Statistical summary of datasets \textbf{SWaT} and \textbf{WADI}.}
\label{tab:wadi}
\resizebox{\linewidth}{!}{%
\begin{tabular}{@{}lcc@{}}
\toprule
Datasets          & SWaT                  & WADI                   \\ \midrule
\textbf{Feature Desc.}           & \multicolumn{2}{c}{All sensors and actuators.} \\
\textbf{\# Features}                   & 51                    & 112                    \\
\textbf{\# Attacks}                    & 41                    & 15                     \\
\textbf{Attack durations (mins)}       & 2 $\sim$ 25            & 1.5 $\sim$ 30           \\
\textbf{Training size (normal data)}   & 49619                 & 120899                 \\
\textbf{Testing size (data with attacks)} & 44931                 & 17219                  \\
\textbf{Anomaly rate (\%)}             & 12.14                 & 5.75                   \\ \bottomrule
\end{tabular}%
}
\end{table}

%----------------------------------------------------------------------------------

\begin{table}[htb]
\caption{Statistical summary of datasets \textbf{SMAP} and \textbf{MSL}.}
\label{tab:nasa}
\resizebox{\linewidth}{!}{%
\begin{tabular}{@{}lcc@{}}
\toprule
Datasets                                    & SMAP   & MSL   \\ \midrule
\textbf{Feature Desc.} & \multicolumn{2}{c}{\begin{tabular}[c]{@{}c@{}}Radiation, temperature, \\ power, etc.\end{tabular}} \\
\textbf{\# Features}                        & 25     & 25    \\
\textbf{Training size (normal data)}        & 135183 & 58317 \\
\textbf{Testing size (data with anomalies)} & 427617 & 73729 \\
\textbf{Anomaly rate (\%)}                  & 13.13  & 10.72 \\ \bottomrule
\end{tabular}%
}
\end{table}

\subsection{Experimental Setup}
\subsubsection{Data preprocessing}
We perform a data standardization before training to improve the robustness of our model. Data preprocessing is applied on both training and testing set:
\begin{equation}
    \tilde{x} = \frac{x-\min X_{train}}{\max X_{train} - \min X_{train}}
\end{equation}
where $\max(X_{train})$ and $\min(X_{train})$ are the maximum value and the minimum value of the training set respectively.

%----------------------------------------------------------------------------------

\begin{table*}[htb]
\caption{Experimental results on \textbf{SWaT} and \textbf{WADI}.}
\label{tab:wadi-res}
\centering
\resizebox{.5\linewidth}{!}{%
\begin{tabular}{@{}clccc@{}}
\toprule
Datasets & Methods    & Precision(\%)      & Recall(\%)         & F1-score            \\ \midrule
         & PCA        & 24.92          & 21.63          & 0.23          \\
         & KNN        & 7.83           & 7.83           & 0.08          \\
         & FB         & 10.17          & 10.17          & 0.10          \\
         & AE         & 72.63          & 52.63          & 0.61          \\
         & DAGMM      & 27.46          & 69.52          & 0.39          \\
         & LSTM-VAE   & 96.24          & 59.91          & 0.74          \\
         & MAD-GAN    & 98.97          & 63.74          & 0.77          \\
         & GDN        & \textbf{99.35} & {\ul 68.12}    & {\ul 0.81}    \\ \cmidrule(l){2-5} 
         & GTA$^{*}$ (ours) & 74.91          & \textbf{96.41} & 0.84 \\
         & GTA$^{**}$ & 94.83 & 88.10 & \textbf{0.91} \\
\multirow{-10}{*}{SWaT} &
  \multicolumn{1}{c}{$\Delta_{\uparrow}$ (best F1)}&
  \multicolumn{1}{c}{\cellcolor[HTML]{C0C0C0}-4.55\%} &
  \multicolumn{1}{c}{\cellcolor[HTML]{C0C0C0}\textbf{+29.33}\%} &
  \multicolumn{1}{c}{\cellcolor[HTML]{C0C0C0}\textbf{+12.35}\%} \\ \midrule
         & PCA        & 39.53          & 5.63           & 0.10          \\
         & KNN        & 7.76           & 7.75           & 0.08          \\
         & FB         & 8.60           & 8.60           & 0.09          \\
         & AE         & 34.35          & 34.35          & 0.34          \\
         & DAGMM      & 54.44          & 26.99          & 0.36          \\
         & LSTM-VAE   & 87.79          & 14.45          & 0.25          \\
         & MAD-GAN    & 41.44          & 33.92          & 0.37          \\
         & GDN        & \textbf{97.50} & {\ul 40.19}    & {\ul 0.57}    \\ \cmidrule(l){2-5} 
         & GTA$^{*}$ (ours) & 74.56          & \textbf{90.50} & 0.82 \\
         & GTA$^{**}$ & 83.91 & 83.61 & \textbf{0.84} \\
\multirow{-10}{*}{WADI} &
  \multicolumn{1}{c}{$\Delta_{\uparrow}$ (best F1)} &
  \multicolumn{1}{c}{\cellcolor[HTML]{C0C0C0}-13.94\%} &
  \multicolumn{1}{c}{\cellcolor[HTML]{C0C0C0}\textbf{+108.04}\%} &
  \multicolumn{1}{c}{\cellcolor[HTML]{C0C0C0}\textbf{+47.37}\%} \\ \bottomrule
  \multicolumn{5}{l}{Best performance in bold. Second-best with underlines.} \\
  \multicolumn{5}{l}{$^{*}$ represents the results chosen by best Recall.} \\
  \multicolumn{5}{l}{$^{**}$ represents the results chosen by best F1-score.} \\
  \multicolumn{5}{l}{$\Delta_{\uparrow}$ represents the percentage increase between our best F1-score} \\
  \multicolumn{5}{l}{performance and the second-best method (GDN).}
\end{tabular}%
}
\end{table*}

%----------------------------------------------------------------------------------

\begin{table*}[htb]
\caption{Experimental results on \textbf{SMAP} and \textbf{MSL}.}
\label{tab:nasa-res}
\centering
\resizebox{.8\linewidth}{!}{%
\begin{tabular}{@{}clcccccc}
\toprule
\multicolumn{2}{c}{}                     & \multicolumn{3}{c}{SMAP} & \multicolumn{3}{c}{MSL}                       \\ \cmidrule(l){3-5} \cmidrule(l){6-8} 
\multicolumn{2}{c}{\multirow{-2}{*}{Method}} &
  Precision(\%) &
  Recall(\%) &
  F1-score &
  Precision(\%) &
  Recall(\%) &
  F1-score \\ \midrule
                           & KitNet      & 77.25  & 83.27  & 0.8014 & 63.12          & 79.36          & 0.7031     \\
                           & GAN-Li      & 67.10  & 87.06  & 0.7579 & 71.02          & 87.06          & 0.7823  \\
                           & LSTM-VAE    & 85.51  & 63.66  & 0.7298 & 52.57          & 95.46          & 0.6780  \\
                           & MAD-GAN     & 80.49  & 82.14  & 0.8131 & 85.17          & 89.91          & 0.8747   \\
\multirow{-5}{*}{R-Models} & OmniAnomaly & 74.16  & \textbf{97.76}  & 0.8434 & {\ul 88.67}          & 91.17          & 0.8989        \\ \cmidrule(l){2-8} 
                           & LSTM-NDT    & \textbf{89.65}  & 88.46  & 0.8905 & 59.44          & 53.74          & 0.5640        \\
                           & DAGMM       & 58.45  & 90.58  & 0.7105 & 54.12          & \textbf{99.34} & 0.7007        \\
                           & MTAD-GAT    & 89.06  & 91.23  & {\ul 0.9013} & 87.54          & 94.40          & {\ul 0.9084}   \\
\multirow{-5}{*}{F-Models}  & GTA$^{**}$ (ours)  &    89.11    & 91.76    & \textbf{0.9041}   & \textbf{91.04} & 91.17          & \textbf{0.9111}     \\

%   \multicolumn{1}{c}{$\Delta_{\uparrow}$} &
%   \cellcolor[HTML]{C0C0C0} & \cellcolor[HTML]{C0C0C0} & \cellcolor[HTML]{C0C0C0} & \cellcolor[HTML]{C0C0C0}2.67\% & \cellcolor[HTML]{C0C0C0} & \cellcolor[HTML]{C0C0C0}1\%
\bottomrule
\multicolumn{8}{l}{Best performance in bold. Second-best with underlines.} \\
\multicolumn{8}{l}{$^{**}$ represents the results chosen by best F1-score.} \\
\end{tabular}%
}
\end{table*}

\subsubsection{Evaluation metrics}
We adopt the standard evaluation metrics in anomaly detection tasks, namely Precision, Recall and F1 score, to evaluate the performance of our approach, in which:
\begin{equation}
    \mathrm{Precision} = \frac{\mathrm{TP}}{\mathrm{TP}+\mathrm{FP}}
\end{equation}
\begin{equation}
    \mathrm{Recall} = \frac{\mathrm{TP}}{\mathrm{TP}+\mathrm{FN}}
\end{equation}
\begin{equation}
    \mathrm{F1} = 2\times \frac{\mathrm{Precision}\times \mathrm{Recall}}{\mathrm{Precision}+\mathrm{Recall}}
\end{equation}
where $\mathrm{TP}$ represents the truly detected anomalies (aka. true positives), $\mathrm{FP}$ stands for the falsely detected anomalies (aka. false positives), $\mathrm{TN}$ represents the correctly classified normal samples (aka. true negatives), and $\mathrm{FN}$ is the misclassified normal samples (aka. false negatives). Given the fact that in many real-world anomaly detection scenarios, it is more vital for the system to detect all the real attacks or anomalies by tolerating a few false alarms. As such, we generally give more concern to $\mathrm{Recall}$ and the overall $\mathrm{F1}$ score instead of $\mathrm{Precision}$. Considering different anomaly score thresholds may result in different metric scores, we hence report both our best $\mathrm{Recall}$ and $\mathrm{F1}$ results (with notations $^{*}$ and $^{**}$ respectively) on all datasets for a thorough comparison.

Also, we adopt the point-adjust way to calculate the performance metrics following \cite{Su2019}. In practice, anomalous observations usually occur consecutively to form contiguous anomaly segments. An alert for anomalies can be triggered within any subset of an actual anomaly window. Thus, for any observation in the ground truth anomaly segment, if it is detected as an anomaly or attack, we would consider this whole anomaly window is correctly detected and every observation point in this segment has been classified as anomalies. The observations outside the ground truth anomaly segment are treated as usual. In all, we first train our model on the training set to learn the general sequence pattern and make the forecasting on the test set for anomaly detection. 

\subsubsection{Baselines}
We compare our GTA with a wide range of state-of-the-arts in multivariate time series anomaly detection, including: (1) reconstruction-based models: PCA, AE \cite{Aggarwal2013}, KitNet \cite{Mirsky2018}, DAGMM \cite{Zong2018}, GAN-Li \cite{Li2018}, OmniAnomaly \cite{Su2019}, LSTM-VAE \cite{Park2017}, MAD-GAN \cite{Li2019}, and (2) forecasting-based models: KNN \cite{Angiulli2002}, FB \cite{Lazarevic2005}, MTAD-GAT \cite{Zhao2020} and GDN \cite{Deng2021}.

\subsubsection{Training Settings}
We implement our method and all its variants using Pytorch\footnote{https://pytorch.org/} version 1.7.0 with CUDA 10.1 and Pytorch Geometric Library \cite{Fey2019} version 1.6.3. We conduct all experiments on four NVIDIA Tesla P100 GPUs. For time series forecasting, we set the historical window size to 60 frames with a label series length as 30 to predict the value at next timestamp. The number of dilated convolution levels for temporal context modeling is set to 3. Also, the general model input embedding dimension is set to 128. For the conventional multi-head attention mechanism, the number of heads is set to 8. In total, we have 3 encoder layers and 2 decoder layers and the dimensional of fully connected network is set to 128 which is equal to the model dimension. Additionally, we apply the dropout strategy to prevent overfitting with dropout rate consistently equals to 0.05. The models are trained using the Adam optimizer with learning rate initialized as $1e^{-4}$ and $\beta_{1}, \beta_{2}$ as 0.9, 0.99, respectively. A learning rate adjusting strategy is also applied. We train our models for up to 50 epochs and early stopping strategy is applied with patience of 10. We run each experiment for 5 trials and report the mean value.

\subsection{Experimental Results}

% \begin{figure}
%     \centering
%     \includegraphics[width=\linewidth]{figs/wadi-case-study.png}
%     \caption{Caption}
%     \label{fig:my_label}
% \end{figure}
In Table \ref{tab:wadi-res}, we show the anomaly detection accuracy in terms of precision, recall, and F1-score, of our proposed GTA method and other state-of-the-arts on datasets \textbf{SWaT} and \textbf{WADI}. Each of these baselines provides a specific threshold selection method, and the reported F1-score is calculated correspondingly. Our proposed GTA significantly outperforms all the other approaches on both datasets by achieving the best F1-score as 0.91 for \textbf{SWaT} and 0.84 for \textbf{WADI}. Astonishingly, compared to the second-best model GDN, GTA can achieve an overall 12.35\% increase and an impressive 47.47\% improvement in terms of the best F1-score on these two datasets, respectively. Moreover, we have the following observations: (1) Compared to the conventional unsupervised approaches such as PCA, KNN, FB, deep learning-based techniques (AE, LSTM-VAE, MAD-GAN, etc.) generally have a better detection performance on both datasets. By adopting the recurrent mechanism (RNN, GRU, LSTM) in modeling long sequences and capturing the temporal context dependencies, the deep learning-based methods demonstrate superiority over the conventional methods. (2) DAGMM \cite{Zong2018} aims to handle multivariate data \emph{without} temporal information, indicating the input data contains only one observation instead of a historical time series window. Hence this approach is not suitable for temporal dependency modeling, which is crucial for multivariate time series anomaly detection. (3) Most existing methods are based on recurrent neural networks to capture temporal dependency, including both reconstruction-based models (LSTM-VAE, OmniAnomaly, MAD-GAN) and forecasting-based models (LSTM-NDT, MTAD-GAT). Of which, LSTM-NDT \cite{Hundman2018} is a deterministic model without leveraging stochastic information for modeling the inherent stochasticity of time series. LSTM-VAE \cite{Park2017} combines LSTM with VAE for sequence modeling; however, it ignores the temporal dependencies among latent variables. OmniAnomaly \cite{Su2019} was then proposed to solve this problem. Additionally, MAD-GAN \cite{Li2019} aims to adopt a general adversarial training fashion to reconstruct the original time series, which also uses recurrent neural networks. Nevertheless, the recurrent learning mechanism's core properties restrict the modeling process to be sequential. Past information has to be retained through the past hidden states, limiting the long-term sequence modeling capability of the model. Transformer adopts a non-sequential learning fashion, and the powerful self-attention mechanism makes the context distance between any token of a time series shrink to one, which is of high importance to sequence modeling as more historical data can provide more pattern information. (4) Though GDN \cite{Deng2021} is also a graph learning-based anomaly detection approach, it adopts the top-K \emph{nearest} connection strategy to model the topological graph structure among sensors, which have certain limitations as we discussed in Section \ref{intro}. MTAD-GAT \cite{Zhao2020} directly utilizes the initial graph structure information by assuming all sensors are mutually connected, making it a complete graph that is not suitable for many real-life situations. 

From Table \ref{tab:nasa-res}, we can see that the overall improvements in terms of best F1-score on datasets \textbf{SMAP} and \textbf{MSL} are not as impressive as Table \ref{tab:wadi-res} shows. We argue the main difference of results between the NASA anomaly datasets and the Cyber-attack datasets lies in the features' dependencies. SMAP provides measurements of the land surface soil moisture by measuring various separated attributes such as radiation, temperature, computational activities, etc. Though these attributes are not entirely independent of each other, the internal relationships between them are much weaker than those within \textbf{SWaT} or \textbf{WADI} where any slight change that appears on one sensor can propagate to the whole network. Therefore, our proposed graph structure learning strategy might be more effective on datasets with a strong topological structure.

\begin{table}
\caption{Anomaly detection accuracy in terms of precision(\%), recall(\%), and F1-score of GTA and its variants.}
\label{tab:wadi-res-2}
\resizebox{\linewidth}{!}{%
\begin{tabular}{@{}lcccccc@{}}
\toprule
\multicolumn{1}{c}{\multirow{2}{*}{Method}} & \multicolumn{3}{c}{SWaT} & \multicolumn{3}{c}{WADI} \\ \cmidrule(l){2-4} \cmidrule(l){5-7} 
\multicolumn{1}{c}{}                        & Prec(\%)    & Rec(\%)    & F1-score    & Prec(\%)    & Rec(\%)    & F1-score    \\ \midrule
GTA                     & \textbf{94.83}   & \textbf{88.10}  & \textbf{0.91}  & \textbf{83.91}   & \textbf{83.61}  & \textbf{0.84}  \\ \midrule
w/o Graph               & 88.64   & 65.73  & 0.75  & 71.25   & 68.23  & 0.70  \\
w/o LP                  & 89.36   & 72.12  & 0.80  & 79.56   & 77.10  & 0.78  \\
w/o Attn                & 78.75   & 65.34  & 0.71  & 74.75   & 70.90  & 0.73  \\ \bottomrule
\end{tabular}%
}
\end{table}

\begin{figure*}[htb]
  \vspace{-1cm}
  \begin{subfigure}[b]{\columnwidth}
    \hspace{.2cm}
    \includegraphics[width=0.9\linewidth]{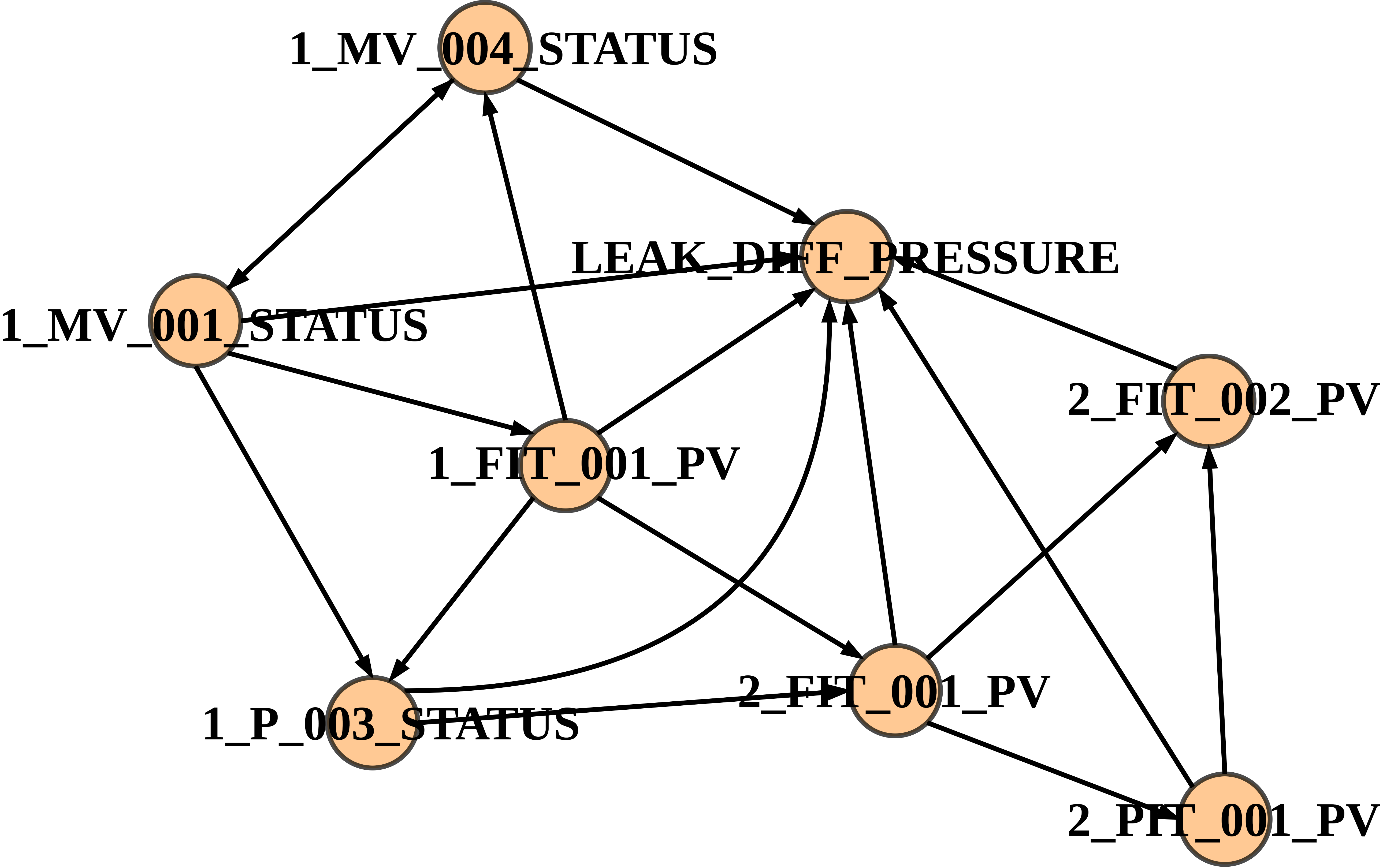}
    % \vspace{-4cm}
    \caption{Partial graph structure learned by the learning policy.}
    \label{fig:case-a}
  \end{subfigure}
  \hfill %%
  \begin{subfigure}[b]{\columnwidth}
    \hspace{-1cm}
    \includegraphics[width=1.2\linewidth]{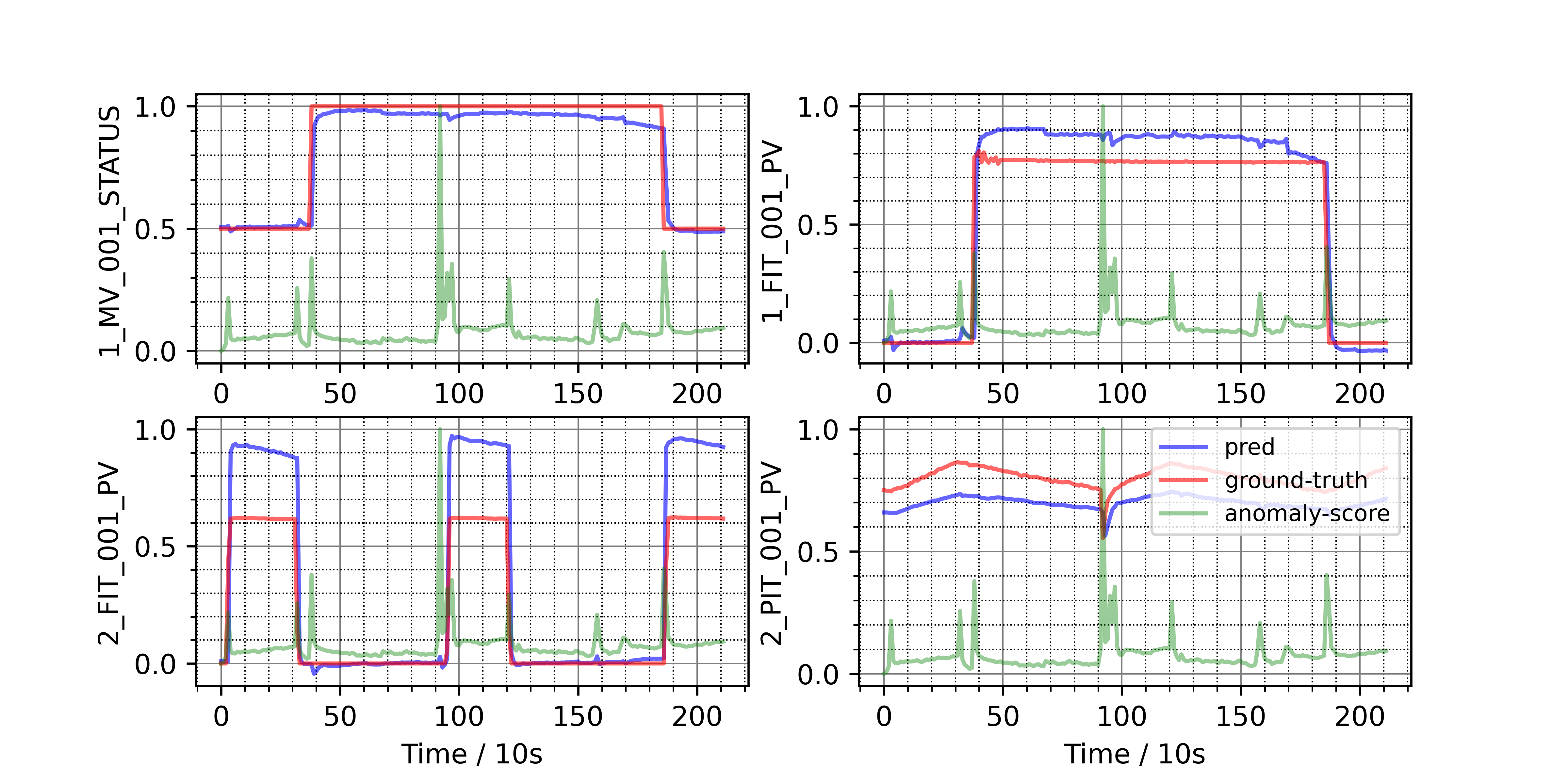}
    \caption{The attacked sensor with three other malicious sensors.}
    \label{fig:case-b}
  \end{subfigure}
  \caption{A case study of showing an attack in \textbf{WADI}.}
%   \label{fig:case}
\end{figure*}

\subsection{Ablation Studies}
To study each component of our approach's effectiveness, we gradually exclude the elements to observe how the model performance degrades on datasets \textbf{SWaT} and \textbf{WADI}. First, we study the significance of modeling the dependencies among sensors using graph learning. We directly apply the original time series as the inputs for the Transformer and make the forecasting without graph learned phase. Second, we study the significance of our proposed \emph{structure learning policy} (LP) by substituting it with a static complete graph where every node is bi-directionally linked to each other. Finally, to study the necessity of the Transformer-based architecture for sequence modeling, we substitute the Transformer with a GRU-based recurrent neural network for forecasting. The results are summarized in Table \ref{tab:wadi-res-2} and provide the following observations: (1) Our proposed learning policy helps the graph convolution operation by capturing only proper information flow with noises filtered out. (2) There is a considerable gap between GTA and the variant without graph learning which again demonstrates the importance of topological structure modeling in handling multivariate time series anomaly detection. (3) Transformer-based architecture exhibits superiority in sequence modeling, where the self-attention mechanism plays a critical role. Moreover, these results again confirm that every component of our method is indispensable and make this framework powerful in multivariate time series anomaly detection. 

\subsection{Graph Learning and Case Study}
By introducing a case study of an actual attack from the Cyber-attack dataset \textbf{WADI}, we evaluate what a graph structure would the graph learning policy learn and how this helps us localize and comprehend an anomaly in this section. An assault with a period of 25.16 minutes was logged in the WADI data collecting log, which fraudulently turned on the motorized valve 1\_MV\_001\_STATUS and caused an overflow on the primary tank. It's difficult for the operation engineers to find out the status of this valve manually because it's still within normal range. As a result, it's not easy to spot this oddity.

The water distribution treatment, for example, consists of three-state processes from the water supply, distribution network, and return water system, which are denoted as P1, P2, and P3, respectively. Every sensor and actuator, in every condition, is inextricably linked. The raw water inlet valve that regulates the SUTD entering, for example, is represented by 1\_MV\_001\_STATUS. Because 1\_FIT\_001\_PV is a downstream flow indicator transmitter of the water distribution, the value of 1\_FIT\_001\_PV rises rapidly if the 1\_MV\_001 valve is turned on abruptly. As its outcome of the first stage propagates the influence from the raw water transfer pump to the second stage, 2\_FIT\_001\_PV is also vulnerable to the same malicious attack. In addition, as LEAK\_DIFF\_PRESSURE becomes irregular during this procedure, the leaking water pressure grows without a doubt. Our graph learning policy learned a partial graph in Fig. \ref{fig:case-a}, which almost properly depicts the topological interactions among sensors. The LEAD\_DIFF\_PRESSURE is almost related to every other displayed node as malicious information passes from upstream sensors to downstream ones. More importantly, Fig. \ref{fig:case-b} shows our model's predicted sensor curves (blue lines) against the ground truth (red lines) of sensor 1\_FIT\_001\_PV, 2\_FIT\_001\_PV, and 2\_PIT\_001\_PV within the attack duration. The predictions of these sensors are consistently higher than the ground truth, where the anomaly score increases correspondingly. It is mainly because the input time series has been embedded with the graph structure information through the influence propagation convolution operation. Sensors that are not directly attacked will still be severely affected if sensors that are highly related to them are attacked. Therefore, our model can capture this dependency and result in an \emph{abnormal} prediction, which is vital for the following anomaly detection.

%----------------------------------------------------------------------------------

% \begin{figure}
%     \centering
%     \includegraphics[width=\linewidth]{figs/main-flow.png}
%     \caption{Caption}
%     \label{fig:main_flow}
% \end{figure}

%----------------------------------------------------------------------------------

% \begin{figure*}[htb]
%     \centering
%     \includegraphics[width=.5\linewidth]{figs/SWaT_pred.png}
%     \caption{Caption}
%     \label{fig:swat_pred}
% \end{figure*}

%----------------------------------------------------------------------------------

% \begin{figure*}[htb]
%     \centering
%     \includegraphics[width=.5\linewidth]{figs/WADI_pred.png}
%     \caption{Caption}
%     \label{fig:my_label}
% \end{figure*}

%----------------------------------------------------------------------------------

%----------------------------------------------------------------------------------

\section{Conclusion}
In this work, we proposed GTA, a Transformer-based framework for anomaly detection that uses the introduced connection learning policy to automatically learn sensor dependencies. To simulate the information flow among the sensors in the graph, we devised an unique Influence Propagation (IP) graph convolution. The inference speed of our proposed multi-branch attention technique is greatly improved without sacrificing model performance. Extensive experiments on four real-world datasets demonstrated that our strategy outperformed other state-of-the-art approaches in terms of prediction accuracy. We also provided a case study to demonstrate how our approach identifies the anomaly by utilizing our proposed techniques. We aim to explore more about combining this approach with the online learning strategy to land it on the mobile IoT scenarios for future work.

% or


% \appendices

% % you can choose not to have a title for an appendix
% % if you want by leaving the argument blank
% \section{}

% \section*{Acknowledgment}


\ifCLASSOPTIONcaptionsoff
  \newpage
\fi



\bibliographystyle{IEEEtran}
\bibliography{reference}
\end{document}